%% file: ijcai24.tex
\documentclass{article}
\usepackage{ijcai24}

\usepackage{times}
\usepackage{soul}
\usepackage{url}
\usepackage[hidelinks]{hyperref}
\usepackage[utf8]{inputenc}
\usepackage[small]{caption}
\usepackage{graphicx}
\usepackage{amsmath}
\usepackage{amsthm}
\usepackage{booktabs}
\usepackage{algorithm}
\usepackage{algorithmic}
\usepackage[switch]{lineno}
\usepackage{subcaption}
\usepackage{xcolor}
\usepackage{amssymb} 
\usepackage{colortbl}
\usepackage{multirow}

\title{ESP-PCT: Enhanced VR Semantic Performance  through Efficient Compression of Temporal and Spatial Redundancies in Point Cloud Transformers}

\author{
Luoyu Mei$^{1,2}$
\and
Shuai Wang$^{1*}$\and
Yun Cheng$^{3}$\thanks{Shuai Wang and Yun Cheng are co-corresponding authors.}\and
Ruofeng Liu$^4$\and
Zhimeng Yin$^2$\and \\
Wenchao Jiang$^5$\and
Shuai Wang$^1$\And
Wei Gong$^6$\\
\affiliations
$^1$Southeast University,\\
$^2$City University of Hong Kong,\\
$^3$Swiss Data Science Center, Zurich, Switzerland,\\
$^4$Robert Bosch LLC,\\
$^5$Singapore University of Technology and Design,\\
$^6$University of Science and Technology of China\\
\emails
lymei-@seu.edu.cn,
shuaiwang\_iot@seu.edu.cn,
yun.cheng@sdsc.ethz.ch,
liux4189@gmail.com,
zhimeyin@cityu.edu.hk,
wenchao\_jiang@sutd.edu.sg,
shuaiwang@seu.edu.cn,
weigong@ustc.edu.cn
}
\def\ShowComment{1} 
\if\ShowComment1
 
\fi

\begin{document}

\maketitle
\begin{abstract}
Semantic recognition is pivotal in virtual reality (VR) applications, enabling immersive and interactive experiences. A promising approach is utilizing millimeter-wave (mmWave) signals to generate point clouds. However, the high computational and memory demands of current mmWave point cloud models hinder their efficiency and reliability. To address this limitation, our paper introduces ESP-PCT, a novel Enhanced Semantic Performance Point Cloud Transformer with a two-stage semantic recognition framework tailored for VR applications. ESP-PCT takes advantage of the accuracy of sensory point cloud data and optimizes the semantic recognition process, where the localization and focus stages are trained jointly in an end-to-end manner. We evaluate ESP-PCT on various VR semantic recognition conditions, demonstrating substantial enhancements in recognition efficiency. Notably, ESP-PCT achieves a remarkable accuracy of 93.2\% while reducing the computational requirements (FLOPs) by 76.9\% and memory usage by 78.2\% compared to the existing Point Transformer model simultaneously. These underscore ESP-PCT's potential in VR semantic recognition by achieving high accuracy and reducing redundancy. The code and data of this project are available at \url{https://github.com/lymei-SEU/ESP-PCT}.
\end{abstract}

\input{sec1_introduction}
\input{sec2_relatedwork}
\input{sec3_ESP-PCT}
\input{sec4_experiments}
\input{sec5_conclusion}
\section*{Acknowledgements}
We sincerely thank the anonymous area chair and reviewers for their valuable comments. This work was supported in part by the National Natural Science Foundation of China under Grant No. 62272098 and the Ministry of Education, Singapore, under its Joint SMU-SUTD Grant (22-SIS-SMU-052).

\bibliographystyle{named}
\bibliography{ijcai24}

\end{document}

%% file: sec1_introduction.tex
\section{Introduction}
Virtual Reality (VR) has experienced rapid growth over the past decade, enhancing user experiences in fields like entertainment, shopping, healthcare, and education~\cite{VRSurvey,MetaverseSurvey}. This evolution is largely driven by advanced sensing capabilities that extract semantic information from VR users, achieved through the recognition and tracking of headset and controller motion. Current VR systems utilize a range of sensors, including Inertial Measurement Units (IMUs)~\cite{VRSurvey} and cameras~\cite{VRCameraSurvey}. Additionally, recent researchers find that integrating millimeter-wave (mmWave) technology~\cite{mmWaveSurvey,li2023egocentric,mmWavemultisurvey} significantly enhances VR sensing capabilities. The mmWave devices, placed in front of the users, produce high-resolution point clouds that accurately depict environments, maintaining fidelity even in obstructions~\cite{mmSpy,Radar2,parsing,cao2022cross}. This approach complements the sensors in VR headsets by providing a third-person perspective. Despite these advantages, employing mmWave radar for precise semantic recognition still presents complex challenges.

Current state-of-the-art designs in this domain are divided into two categories: (i) Vision transformer (ViTs) based methods have outstanding accuracy in processing high-resolution imagery and video~\cite{A_Survey_on_Vision_Transformer,hu2024lf}, but suffer from high computational and storage cost, privacy concerns, and limited perception~\cite{NLOST,Point_Cloud_Matching}. (ii) Point transformer-based methods present effectiveness in handling the sparsity and instability of mmWave point cloud data~\cite{PointTransformer,Pointtransformerv3}, yet facing challenges in focusing on key motion features, reducing model cost, and enhancing robustness against environmental noise~\cite{SuperpointTransformer,Feng_Quan_Wang_Wang_Yang_2024}.

These limitations hinder the widespread utilization of these approaches for VR applications, where real-time processing and responsiveness are crucial for user experience and immersion, as they demand extensive computational resources and struggle to adapt to various environmental conditions. The existing models process entire mmWave point cloud data without prioritizing the semantically relevant information, which is crucial for VR tasks~\cite{End-to-End}. Additionally, these models lead to unnecessary computational overhead, memory waste, and a potential decline in performance efficiency, especially in real-time applications where rapid processing and decision-making are essential. Therefore, an efficient learning framework that localizes and extracts semantic information from the most relevant point cloud data is urgently needed to enhance VR semantic recognition tasks.

To overcome these limitations, we introduce ESP-PCT, a framework designed to optimize the utilization of mmWave point cloud data in VR applications. ESP-PCT tackles two key challenges: (i) How to focus on the moving parts of targets, especially the semantic-discriminative regions, in sparse point clouds, and (ii) How to leverage the point cloud data of these critical parts for enhanced VR semantic recognition. The ESP-PCT model addresses these challenges with a two-stage framework that first localizes key areas (e.g., VR controller) through \textit{Localization Stage}, and then applies attention mechanisms to these selected points in \textit{Focus Stage}. We discover that not all points in the point cloud contribute equally to improving accuracy, while some distract the model. Inspired by this discovery, ESP-PCT concentrates only on the point clouds of the controller, which exhibit denser reflected point clouds. Hence, our two-stage framework significantly narrows the focus to key regions on the VR controllers that positively influence accuracy, drastically reducing computational costs in subsequent stages while eliminating noise from non-essential areas, thus enhancing model accuracy.

Specifically, based on a point transformer architecture~\cite{PointTransformer}, ESP-PCT employs the localization stage that analyzes the raw point cloud data to make early identification. This stage processes data efficiently and utilizes smart strategies to reuse features, which saves computational resources. This consistency is critical for smooth training from start to finish, beneficial for saving resources, and keeping critical contextual details for the focus stage. Applying ESP-PCT for VR semantic recognition tasks leads to remarkable results. The ESP-PCT achieves a 93.2\% accuracy while reducing the computational cost, cutting the FLOPs by 76.9\% and memory utilization by 78.2\%, setting new efficiency and performance in VR semantic recognition.

\begin{figure}[t]
    \centering
    \includegraphics[width=1\linewidth]{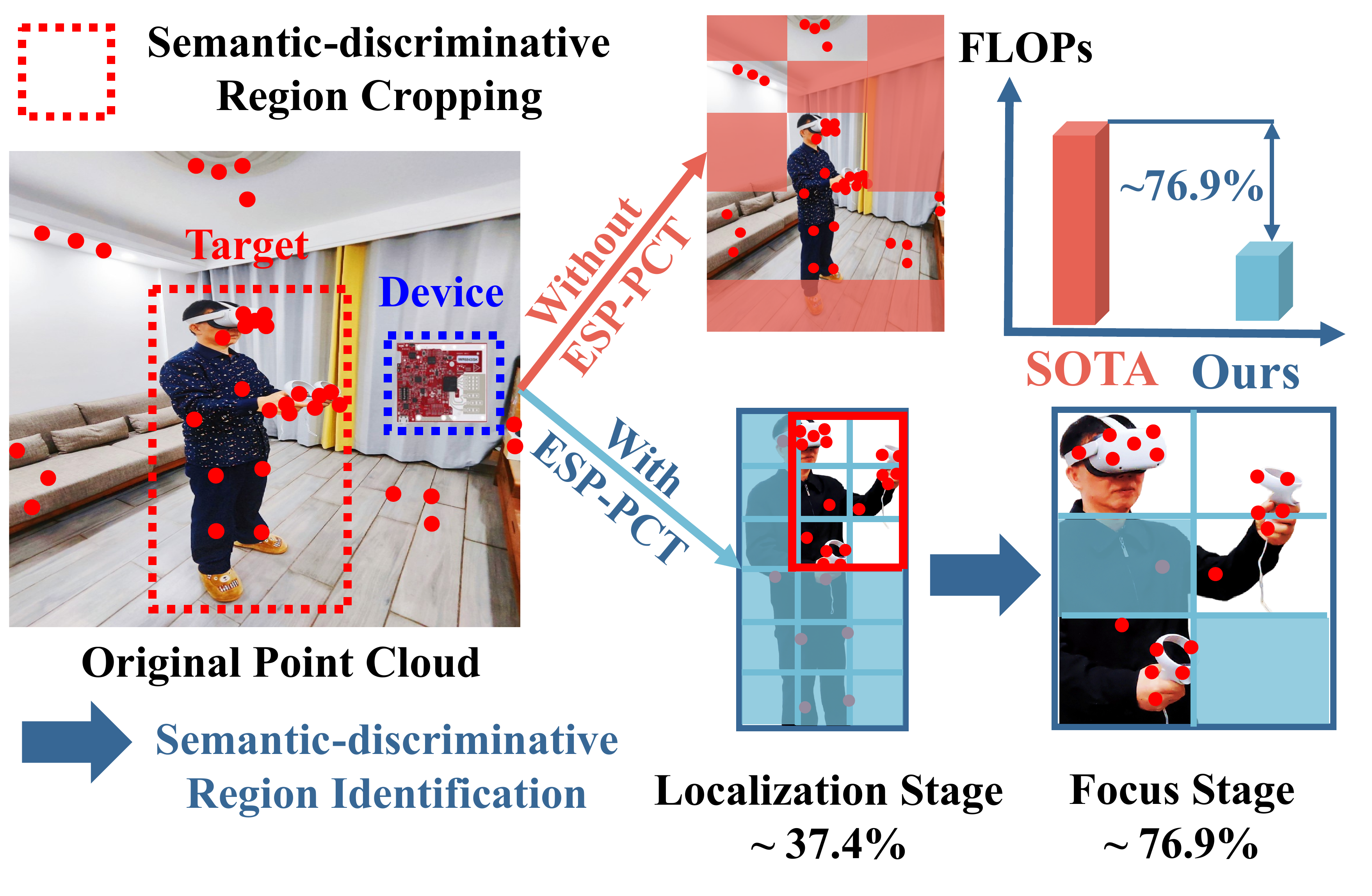}
    \caption{ESP-PCT enhances model accuracy while reducing the model's computational complexity (FLOPs) by 76.9\%. }
    \label{fig:example}
\end{figure}


ESP-PCT is a flexible and robust framework designed for various sub-tasks in semantic recognition. Its adaptability stems from its two-stage structure, enabling it to be reused efficiently across scenarios. The localization and focus stages of ESP-PCT, as outlined in Fig.~\ref{fig:example}, are not just task-specific but are flexible enough to be applied to a range of semantic recognition sub-tasks in VR environments, as shown in the experiments. This reusability is a significant advantage, especially in diverse VR applications.

To summarize, our contribution is three-fold:

\begin{itemize}
\item Introduction of ESP-PCT: this paper introduces ESP-PCT, a novel and efficient two-stage semantic recognition framework tailored for virtual reality (VR) applications. It leverages sparse point cloud data and is designed to optimize both accuracy and computational resources in VR semantic recognition tasks.
\item Flexibility and Reusability: ESP-PCT stands out for its versatility and reusability across a diverse range of VR semantic recognition sub-tasks. Its adaptability enables effective application in various scenarios, making it a highly valuable tool in the evolving domain of VR.
\item Significant Efficiency Improvements: a key achievement of ESP-PCT is its substantial enhancement in computational efficiency. The framework reduces the computational load, reduces FLOPs by 76.9\%, and decreases memory usage by 78.2\%, thereby setting new benchmarks for efficiency in VR semantic recognition.
\end{itemize}

%% file: sec2_relatedwork.tex
\section{Related Work}
\label{sec2_relatedwork}
Existing methodologies in this domain are divided into two categories: Vision and Point Transformer.

\subsection{Vision Transformer}
Vision transformer~\cite{ViViT,CSWinTransformer} is a series of pioneering works that apply the transformer model to image classification by splitting images into patches and treating them as tokens. Vision transformers have achieved competitive performance compared to various vision tasks, such as object detection~\cite{DiffusionDet,ObjectViT}, semantic segmentation~\cite{SemanticViT,SegViT}, and video understanding~\cite{MeMViT,SegViT,RecurringViT}.  However, ViTs are not designed to handle 3D point cloud data, which is irregular, unordered, and sparse~\cite{VideoTransformersSurvey,Point_Cloud_Matching} and lack of effectiveness under non-line-of-sight scenarios~\cite{NLOST,Point_Cloud_Matching}. These limitations make vision transformer-based methods unsuitable for our VR semantic recognition scenario, which requires efficient and robust performance in various environments~\cite{Multimodal_Learning_With_Transformers_A_Survey,A_Survey_on_Vision_Transformer}.

\begin{figure*}[t]
    \centering
    \includegraphics[width=1\linewidth]{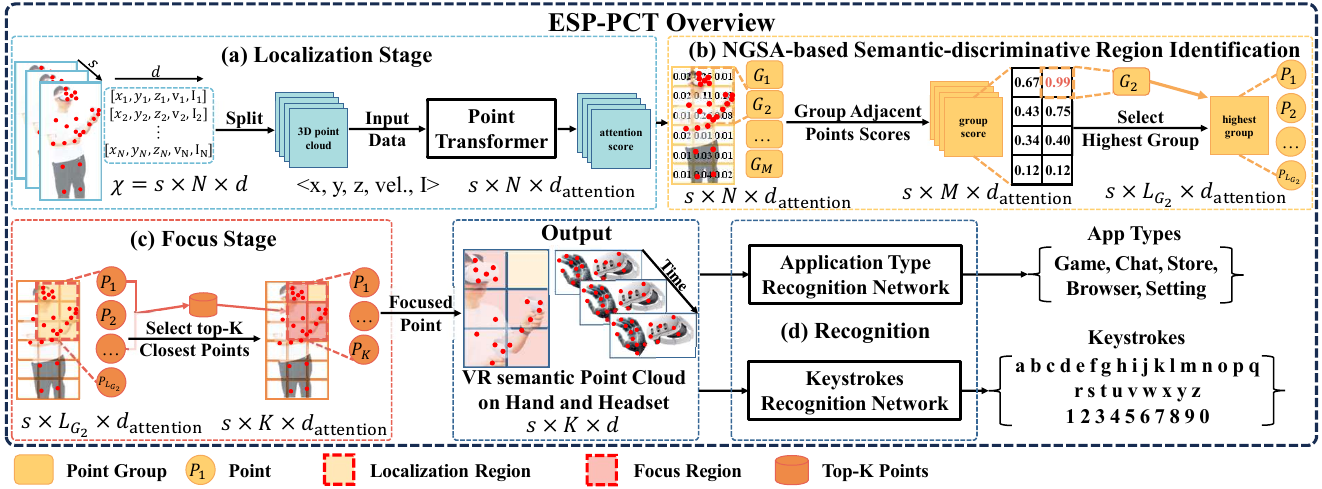}
    \caption{Overview of ESP-PCT. (a) ESP-PCT inputs the point cloud into a point transformer, which assigns attention scores for each point. (b) The adjacent points in the point cloud are grouped, and a collective group score is computed. When semantic resolution is unclear, the Neighborhood Global Semantic Attention (NGSA) mechanism searches semantically discriminative regions. (c) The top-K points with the highest global semantic attention are chosen for focused recognition, with the efficiency of this stage bolstered by feature fusion strategies. (d) The recognition networks take the VR semantic point cloud as input and perform further semantic recognition.}
    \label{fig:LFoverview}
\end{figure*}

\subsection{Point Transformer}
Point transformers~\cite{PointTransformer,PointTransformerV2,3D_Point_Clouds_Transformer} is a family of neural networks that apply the transformer model to point clouds without voxelization or graph construction. Stratified point transformer~\cite{StratifiedPCT} utilizes a stratified transformer layer to capture the hierarchical structure and feature fusion of point clouds. Point 4D transformer~\cite{Point4DTransformer} extends the point transformer model to the 4D space-time domain by adding a temporal attention layer and a spatiotemporal fusion layer to model the dynamics and correlations of point cloud videos. Self-supervised 4D~\cite{Self-Supervised4D} develops a self-supervised learning framework for point cloud video representation learning by distilling and reconstructing the point cloud sequence. Interpretable3D~\cite{Feng_Quan_Wang_Wang_Yang_2024} prioritizes interpretability in dense point clouds. However, they still face challenges in focusing on reducing computational cost and enhancing robustness against environmental noise. Unlike the existing approaches, ESP-PCT aims to improve the efficiency of the point transformer for VR semantics. ESP-PCT effectively identifies semantic recognition in sparse point clouds, focusing on reducing computational load and memory demands. The detailed design of ESP-PCT is demonstrated in the next Section. 

%% file: sec3_ESP-PCT.tex
\section{ESP-PCT}
\label{sec3_ESP-PCT}
This section illustrates the point transformer model as preliminaries, followed by the two-stage design of ESP-PCT.

\subsection{Preliminaries}
The point transformer model, referenced in~\cite{PointTransformer}, processes multi-frame millimeter-wave point clouds to output attention scores for each point. The input to this neural network is a tensor sized $s \times N \times d$, where $s$ is the combined batch and length size, $N$ is the number of points per frame, and $d$ is the dimension of each point, encompassing five features: $x$, $y$, $z$, velocity, and intensity. The output is a tensor of size $s \times N \times d_{attention}$, with $d_{attention}$ representing the dimension of the output feature vector for each point. The model consists of several layers that apply vector attention to input points. Vector attention in each layer is computed as:

\begin{equation}
\label{eq:PTvector}
\vec{y}_i = \sum{\vec{x}_{j} \in \mathcal{X}} \vec{a}_{ij} \odot \alpha(\vec{x}_{j}),
\end{equation}

\noindent where $\vec{y}_i$ is the output feature vector for the $i$-th point, and $\vec{a}_{ij}$ is the attention weight between points $i$ and $j$. The attention weight $\vec{a}_{ij}$ is calculated utilizing feature transformations and a non-linear function:

\begin{equation}
\label{eq:PTattention}
\vec{a}_{ij} = \rho \left(\gamma \left(\beta \left(\varphi(\vec{x}_{i}), \psi(\vec{x}_{j})\right)+\delta \right)\right),
\end{equation}

\noindent where $\varphi$ and $\psi$ are feature transformations, $\beta$ is a relation function, $\gamma$ a mapping function (usually an MLP), $\delta$ a learned bias, and $\rho$ a non-linear activation function.

\subsection{Localization Stage}
In the ESP-PCT localization stage, as depicted in Fig.~\ref{fig:LFoverview}, we analyze VR user body-generated point cloud data to pinpoint key semantic-discriminative regions. Utilizing a vector attention mechanism specified in Equation~\ref{eq:PTvector}, we compute attention scores for each point, which are crucial for VR semantic recognition and are based on space and feature relations derived from Equation~\ref{eq:PTattention} and allow ESP-PCT to identify the most informative points for detailed scene analysis while ensuring computational efficiency. The model aggregates the attention scores of points that are in close spatial proximity, which is reflected in the following group score equation:

\begin{equation}
\label{eq:group_score_combined}
G_n = \sum_{\vec{a}_{ij} \in G_k} \vec{a}_{ij},
\end{equation}

\noindent where $G_n$ represents the $n$-th group of points. Each group's cumulative score represents the Neighborhood Global Semantic Attention (NGSA). 

The group with the highest NGSA score, highlighted in the point cloud, is deemed the principal semantic-discriminative region. This aggregation process emphasizes collectively significant point cloud regions, enhancing the model's ability to discern semantic-discriminative regions within the cloud, which are crucial for refined semantic recognition tasks. ESP-PCT's selective attention mechanism effectively differentiates between various VR semantics. If the score is lower, we continue to localize semantic-discriminative regions for more precise recognition. This method ensures optimized performance and computational efficiency in our framework. 

\subsection{Semantic-discriminative Regions Identification}
\begin{figure}[t]
    \centering
    \includegraphics[width=1\linewidth]{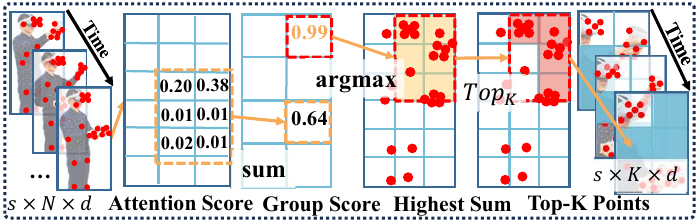}
    \caption{Illustration of ESP-PCT semantic-discriminative region identification. The black numbers indicate the global semantic attention of points. The yellow number indicates the region with the maximum neighborhood global semantic attention (NGSA), which is selected as the VR semantic region.}
    \label{fig:LFPointTransformer}
\end{figure}

As depicted in Fig.~\ref{fig:LFPointTransformer}, we introduce a novel mechanism tailored to address the inherent irregularity and disorder in point cloud data. This mechanism's central innovation is the vector attention mechanism, which offers a distinct approach from the scalar attention weights used in vision transformers~\cite{Transformers_in_Vision_A_Survey}. Vector attention operates at the individual feature channel level, providing a significant advantage in handling the unstructured nature of point cloud data. The semantic-discriminative region identification is beneficial given the complexity of relationships between points in point clouds, as opposed to images' more straightforward pixel grid structure.

In ESP-PCT, the vector attention mechanism is implemented as described in the preliminaries section, with a key modification: we utilize different feature transformations $\varphi(\vec{x}_i)$ and $\psi(\vec{x}_j)$ for the $i$-th and $j$-th points, respectively. These transformations extract the feature representations from each point, which are then processed through the relation function $\beta$, the mapping function $\gamma$, and the non-linear activation function $\rho$, as outlined in Equation~\ref{eq:PTattention}. Additionally, we incorporate a learned bias term $\delta$ before applying the non-linear activation function $\rho$. This bias adds an offset, further refining the effectiveness of the attention mechanism.

After obtaining the vector attention scores for each point in the point cloud, we group the points based on their physical proximity and assign each group a global attention score that reflects its relevance to the target semantic. The global attention score for each group is computed as follows:

\begin{equation}
\label{eq:NGSAglobal}
g_j = \frac{1}{|G_j|} \sum_{\vec{x}_i \in G_j} \vec{y}_i^\intercal \vec{w},
\end{equation}

\noindent where $G_j$ is the $j$-th group of points, $\vec{y}_i$ is the vector attention score for point $\vec{x}_i$, and $\vec{w}$ is a learnable weight vector.

To identify the semantic-discriminative regions, we propose a novel NGSA mechanism, which selects the points with the highest global attention scores and, thus, is most informative for the semantic recognition task. Our novel NGSA mechanism identifies the semantic-discriminative region by selecting the group with the highest $g_j$:

\begin{equation}
\label{eq:NGSAMechanism}
R_S = \underset{j}{\mathrm{argmax}} \, g_j,
\end{equation}

\noindent where \( R_S \) stands for the semantic-discriminative Region. This equation selects the group \( g_j \) that maximizes the global attention score, which is the region with the highest relevance to the target semantic.

\subsection{Focus Stage}

When ESP-PCT's localization stage predicts a result $R_S < \eta$, which is the decision boundary for determining whether to proceed with further localization and focused recognition, triggering the focus stage to refine the process further, this stage involves identifying class-discriminative regions utilizing the NGSA mechanism, as defined in Equation~\ref{eq:NGSAMechanism}. The NGSA mechanism selects the top-K points with the highest global attention scores from the set \( G \), which includes all possible points \( \{g_1, g_2, ..., g_M\} \), where \( M \) is the total number of points. Each group has a corresponding representation \( \vec{h}_k \). The representation \( \vec{z} \) of the point cloud, which is essential for semantic recognition, is constructed by concatenating the representations from the top-K points. This process of concatenation is outlined as follows:

\begin{equation}
\label{eq:NGSAselect}
\vec{z} = \text{Concat}(\{\vec{h}_i | g_i \in \text{Top-K}(G)\}),
\end{equation}

\noindent where $\vec{h}_i$ represents each group's representation and $M$ is the total number of points. The top-K points are selected based on their high scores, as formally expressed by:

\begin{equation}
\text{Top-K}(G) = \{ g_i \in G \,|\, \exists K^{\prime} \subseteq G\},
\end{equation}

\noindent where $|K^{\prime}|=K$ and for all $g_j \in K^{\prime}, s(g_i) \geq s(g_j)$, and for all $g_j \in G \setminus K^{\prime}, s(g_i) > s(g_j)$. Equation~\ref{eq:NGSAselect} is rewritten as:

\begin{equation}
\vec{z} = \text{Concat}\left(\left\{\vec{h}_i \,|\, g_i \in G \text{ and } I_{\text{top-K}}(g_i) = 1\right\}\right).
\end{equation}

This NGSA mechanism extracts the most distinctive regions for each semantic category, effectively filtering out irrelevant or noisy portions of the point cloud. This process extracts the semantic-discriminative region of the model's attention. Extracting this region enhances the model's interpretability and precision in decision-making, as these regions significantly influence the subsequent analysis and recognition stages. The vector attention mechanism thus provides a highly efficient method for aggregating features from the point cloud, capturing complex patterns and relationships. This flexibility positions ESP-PCT as a powerful tool for various semantic recognition tasks, including segmentation and object detection via mmWave point cloud analysis.

Our novel NGSA mechanism can significantly reduce the computation overhead.
The self-attention component within the NGSA mechanism is $O(N^2*D)$, where $N$ represents the input length and $D$ represents the hidden dimension. NGSA employs a minimal number of layers without utilizing multi-head attention, thus significantly reducing costs. This design choice reduces the input length for subsequent multi-head models from $N$ to $K$ (e.g., from 100 to 30) with minimal overhead. Since subsequent functional models typically exhibit a $O(N^2)$ complexity, our approach substantially lowers computational costs.

\subsection{VR Semantic Recognition}
\label{sec_VRSemanticRecognition}

To demonstrate the flexibility and reusability of our approach, we detail the models used for VR semantic recognition in this section, which aim to interpret user intentions and behaviors within VR environments~\cite{VRSPY1}. Therefore, recognizing application types and keystrokes is pivotal in VR semantic recognition~\cite{VRSurvey}. We introduce AppNet and KeyNet, the dedicated models for these two tasks. Specifically, the ESP-PCT preprocess the data before inputting it into these two application domain models and tests the effectiveness of our method.

\begin{figure}[h]
    \centering
    \includegraphics[width=1\linewidth]{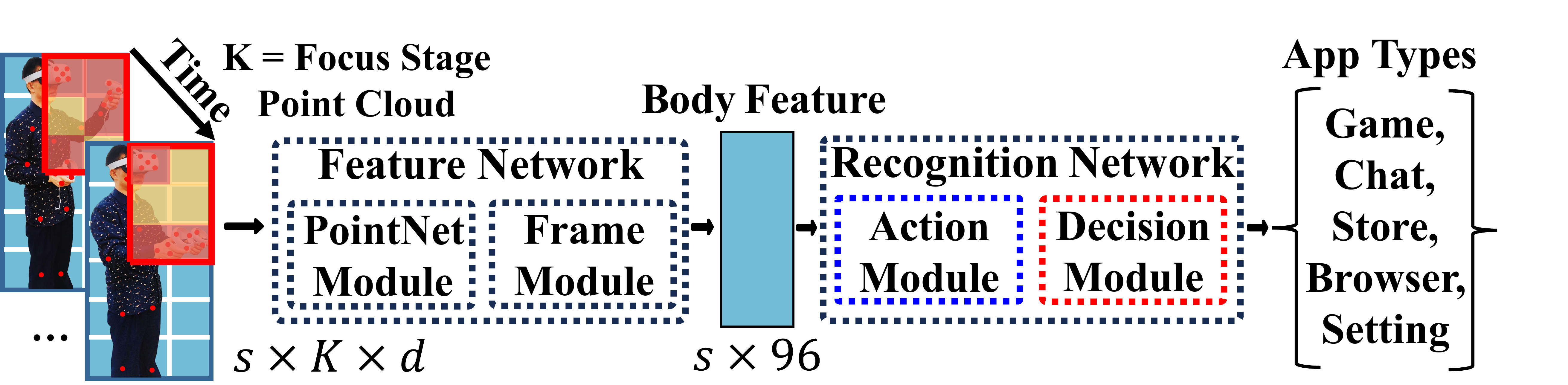}
    \caption{The architecture of AppNet for application recognition.}
    \label{fig:appnet}
\end{figure}

Fig.~\ref{fig:appnet} illustrates the AppNet workflow, which employs point clouds of dimension $ s \times K \times d $, gathered during the focus stage to discern the application type.  The feature network extracts body feature dimensions of $s \times 96$. Then, these body features are sent into the LSTM action module to extract continuous VR actions. Finally, the decision module classifies the output of the action module into five categories of applications. For instance, gaming applications typically involve more body movement, whereas browsing is characterized by hand movements~\cite{VRGit}.

\begin{figure}[h]
    \centering
    \includegraphics[width=1\linewidth]{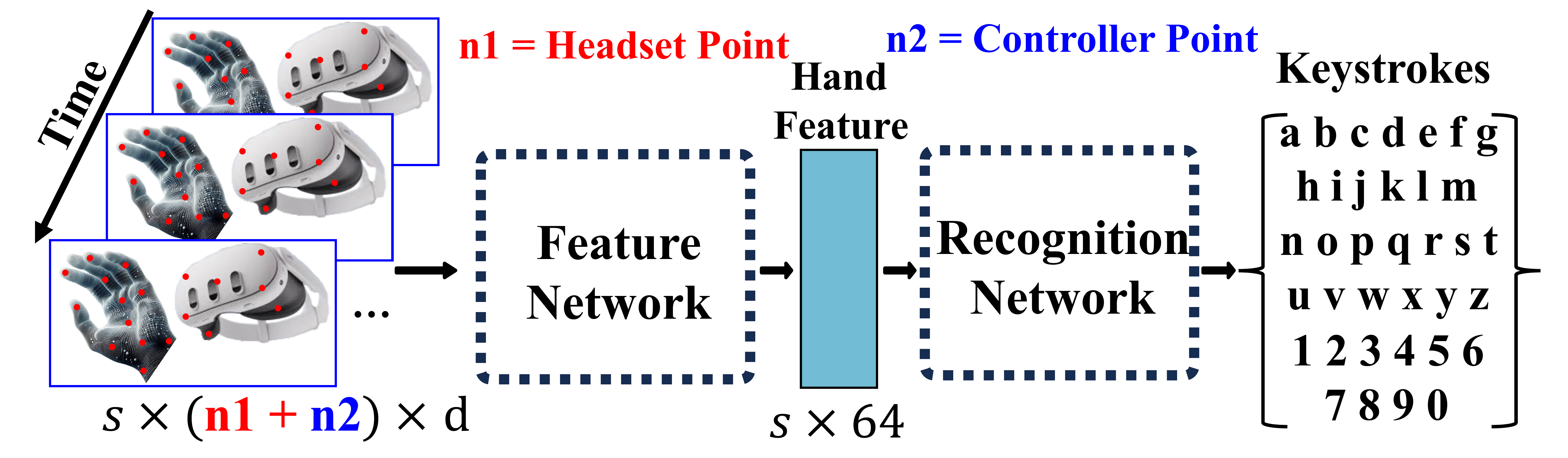}
    \caption{The architecture of KeyNet for keystroke recognition.}
    \label{fig:keynet}
\end{figure}

As illustrated in Fig.~\ref{fig:keynet}, KeyNet is tailored to detect keystrokes from mmWave point clouds collected from a VR user's hands and headset with dimension $s \times (n_1+n_2) \times d$, where $n_1$ and $n_2$ are point cloud on headset and controller. The feature network extracts hand feature dimensions of $s \times 64$. Then, these hand features are sent into the recognition network made of Bi-LSTM~\cite{bilstm}. This model is adept at recognizing contextually related timing features and accurately identifying the keystroke inputs.

%% file: sec4_experiments.tex
\section{Experiments}
\label{sec4_experiments}
\begin{table*}[t]
\centering
\resizebox{1\textwidth}{!}{
    \begin{tabular}{clcccccc}
    \toprule
    Occlusion & Model & \multicolumn{3}{c}{VR Application Type Recognition} & \multicolumn{3}{c}{VR Keystrokes Recognition} \\ 
    \cmidrule(lr){3-5} \cmidrule(lr){6-8}
     &  & Top-1 Acc.(\%) & FLOPs(G) & Params (K) & Top-1 Acc.(\%) & FLOPs(G) & Params (K) \\
    \midrule
    \multirow{8}{*}{\rotatebox[origin=c]{90}{No Occlusion}}
    & Baseline (Attention mechanism) & 81.9 & 4.3 & 2680 & 69.2 & 1.3 & 1285 \\ 
    & Point Transformer~\cite{PointTransformer} & 87.2 & 2.6 & 1680 & 78.7 & 0.8 & 512 \\
    & Point 4D Transformer~\cite{Point4DTransformer} & 91.5 & 2.3 & 1328 & 81.2 & 0.7 & 458 \\
    & Stratified Transformer~\cite{StratifiedPCT} & 92.6 & 1.6 & 1106 & 82.6 & 0.6 & 324 \\
    & Self-Supervised4D~\cite{Self-Supervised4D} & 93.7 & 1.8 & 1239 & 85.4 & 0.6 & 368 \\
    & \textbf{ESP-PCT} (k=32, $\eta$ = 0.45) & \textbf{93.2} & \underline{\textbf{0.6}} & \underline{\textbf{367}} & \textbf{82.1} & \underline{\textbf{0.2}} & \textbf{186} \\
    & \textbf{ESP-PCT} (k=64, $\eta$ = 0.68) & \textbf{95.8} & \textbf{0.9} & \textbf{693} & \textbf{85.3} & \textbf{0.3} & \textbf{215} \\
    & \cellcolor{lightgray}\textbf{ESP-PCT (k=96, $\eta$ = 0.82)} & \cellcolor{lightgray}\textbf{97.6} & \cellcolor{lightgray}\textbf{1.5} & \cellcolor{lightgray}\textbf{986} & \cellcolor{lightgray}\textbf{92.8} & \cellcolor{lightgray}\textbf{0.5}  & \cellcolor{lightgray}\textbf{297} \\ 
    \midrule
    \multirow{8}{*}{\rotatebox[origin=c]{90}{Wood Occlusion}}
    & Baseline (Attention mechanism) & 75.4 & 4.1 & 2598 & 63.1 & 1.2 & 1263 \\
    & Point Transformer~\cite{PointTransformer} & 82.1 & 2.7 & 1631 & 73.4 & 0.9 & 510 \\
    & Point 4D Transformer~\cite{Point4DTransformer} & 86.7 & 2.3 & 1281 & 76.3 & 0.7 & 454 \\
    & Stratified Transformer~\cite{StratifiedPCT} & 88.1 & 1.6 & 1031 & 77.9 & 0.7 & 321 \\
    & Self-Supervised4D~\cite{Self-Supervised4D} & 89.4 & 1.8 & 1123 & 80.8 & 0.5 & 365 \\
    & \textbf{ESP-PCT} (k=32, $\eta$ = 0.45) & \textbf{88.9} & \underline{\textbf{0.4}} & \underline{\textbf{361}} & \textbf{77.6} & \underline{\textbf{0.2}} & \underline{\textbf{183}} \\
    & \textbf{ESP-PCT} (k=64, $\eta$ = 0.68) & \textbf{91.6} & \textbf{0.8} & \textbf{679} & \textbf{80.9} & \textbf{0.3} & \textbf{208} \\
    & \cellcolor{lightgray}\textbf{ESP-PCT} (k=96, $\eta$ = 0.82) & \cellcolor{lightgray}\textbf{94.1} & \cellcolor{lightgray}\textbf{1.6} & \cellcolor{lightgray}\textbf{975} & \cellcolor{lightgray}\textbf{88.7} & \cellcolor{lightgray}\textbf{0.7} & \cellcolor{lightgray}\textbf{293} \\
    \midrule
    \multirow{8}{*}{\rotatebox[origin=c]{90}{Brick Occlusion}}
    & Baseline (Attention mechanism) & 72.8 & 4.2 & 2478 & 60.5 & 1.5 & 1183 \\
    & Point Transformer~\cite{PointTransformer} & 79.6 & 2.7 & 1482& 70.2 & 1.0 & 498 \\
    & Point 4D Transformer~\cite{Point4DTransformer} & 84.4 & 2.4 & 1254 & 73.5 & 0.8 & 433 \\
    & Stratified Transformer~\cite{StratifiedPCT} & 85.9 & 1.8 & 1012 & 75.2 & 0.7 & 209 \\
    & Self-Supervised4D~\cite{Self-Supervised4D} & 87.2 & 1.9 & 1078 & 78.1 & 0.7 & 354 \\
    & \textbf{ESP-PCT} (k=32, $\eta$ = 0.45) & \textbf{86.7} & \underline{\textbf{0.5}} & \underline{\textbf{358}} & \textbf{75.4} & \underline{\textbf{0.2}} & \underline{\textbf{179}} \\
    & \textbf{ESP-PCT} (k=64, $\eta$ = 0.68) & \textbf{89.5} & \textbf{0.9} & \textbf{683} & \textbf{78.6} & \textbf{0.3} & \textbf{201} \\
    & \cellcolor{lightgray}\textbf{ESP-PCT} (k=96, $\eta$ = 0.82) & \cellcolor{lightgray}\textbf{92.3} & \cellcolor{lightgray}\textbf{1.4} & \cellcolor{lightgray}\textbf{963} & \cellcolor{lightgray}\textbf{86.9} & \cellcolor{lightgray}\textbf{0.6} & \cellcolor{lightgray}\textbf{281} \\
    \midrule
    \multirow{8}{*}{\rotatebox[origin=c]{90}{Combined Occlusion}}
    & Baseline (Attention mechanism) & 68.3 & 3.9 & 2318 & 55.7 & 1.1 & 1123 \\
    & Point Transformer~\cite{PointTransformer} & 75.9 & 2.5 & 1287 & 66.8 & 0.9 & 456 \\
    & Point 4D Transformer~\cite{Point4DTransformer} & 80.8 & 2.3 & 1175 & 69.4 & 0.9 & 348 \\
    & Stratified Transformer~\cite{StratifiedPCT} & 82.3 & 1.4 & 974 & 71.1 & 0.7 & 298 \\
    & Self-Supervised4D~\cite{Self-Supervised4D} & 83.6 & 1.5 & 883 & 74.3 & 0.7 & 313 \\
    & \textbf{ESP-PCT} (k=32, $\eta$ = 0.45) & \textbf{83.1} & \underline{\textbf{0.4}} & \underline{\textbf{339}} & \textbf{71.8} & \underline{\textbf{0.1}} & \underline{\textbf{172}} \\
    & \textbf{ESP-PCT} (k=64, $\eta$ = 0.68) & \textbf{86.2} & \textbf{0.7} & \textbf{675} & \textbf{74.7} & \textbf{0.2} & \textbf{198} \\
    & \cellcolor{lightgray}\textbf{ESP-PCT} (k=96, $\eta$ = 0.82) & \cellcolor{lightgray}\textbf{89.4} & \cellcolor{lightgray}\textbf{1.3} & \cellcolor{lightgray}\textbf{957} & \cellcolor{lightgray}\textbf{83.2} & \cellcolor{lightgray}\textbf{0.3} & \cellcolor{lightgray}\textbf{276} \\
    \bottomrule
    \end{tabular}
}
\caption{ESP-PCT is compared with baseline methods under four distinct scenarios, including no occlusion, wood occlusion, brick occlusion, and combined occlusion. Gray cells highlight the models with the highest accuracy, while underlined marks the most efficient models.}
\label{tab:overallperformance}
\end{table*}
\begin{table*}[t]
\centering
\resizebox{1\textwidth}{!}{
    \begin{tabular}{lcccccc}
    \toprule
    Model & \multicolumn{3}{c}{VR Application Type Recognition} & \multicolumn{3}{c}{VR Keystrokes Recognition} \\ 
    \cmidrule(lr){2-4} \cmidrule(lr){5-7}
     & Top-1 Acc.(\%) & FLOPs(G) & Params (K) & Top-1 Acc.(\%) & FLOPs(G) & Params (K) \\ 
    \midrule
    ESP-PCT (w/o calculate the attention score) & 68.7 & 6.2 & 3720 & 48.1 & 1.8 & 2105 \\ 
    ESP-PCT (w/o group the attention score) & 86.3 & 4.6 & 2819 & 77.6 & 1.6 & 1874 \\ 
    ESP-PCT (w/o identify highest score group) & 93.1 & 2.7 & 1803 & 82.5 & 1.2 & 1063 \\ 
    ESP-PCT (w/o select the top-K closest points) & 93.9 & 1.8 & 1248 & 85.3 & 0.7 & 487 \\  
    \cellcolor{lightgray}\textbf{ESP-PCT (k=96, $\eta$ = 0.82)} & \cellcolor{lightgray}\textbf{97.6} & \cellcolor{lightgray}\textbf{1.5} & \cellcolor{lightgray}\textbf{986} & \cellcolor{lightgray}\textbf{92.8} & \cellcolor{lightgray}\textbf{0.5}  & \cellcolor{lightgray}\textbf{297} \\ 
    \bottomrule
    \end{tabular}
}
\caption{Performance of ablation study.}
\label{tab:ablationperformance}
\end{table*}

\subsection{Datasets}
\begin{figure}[t]
\centering
    \begin{subfigure}[t]{0.46\linewidth}
        \includegraphics[width=\linewidth]{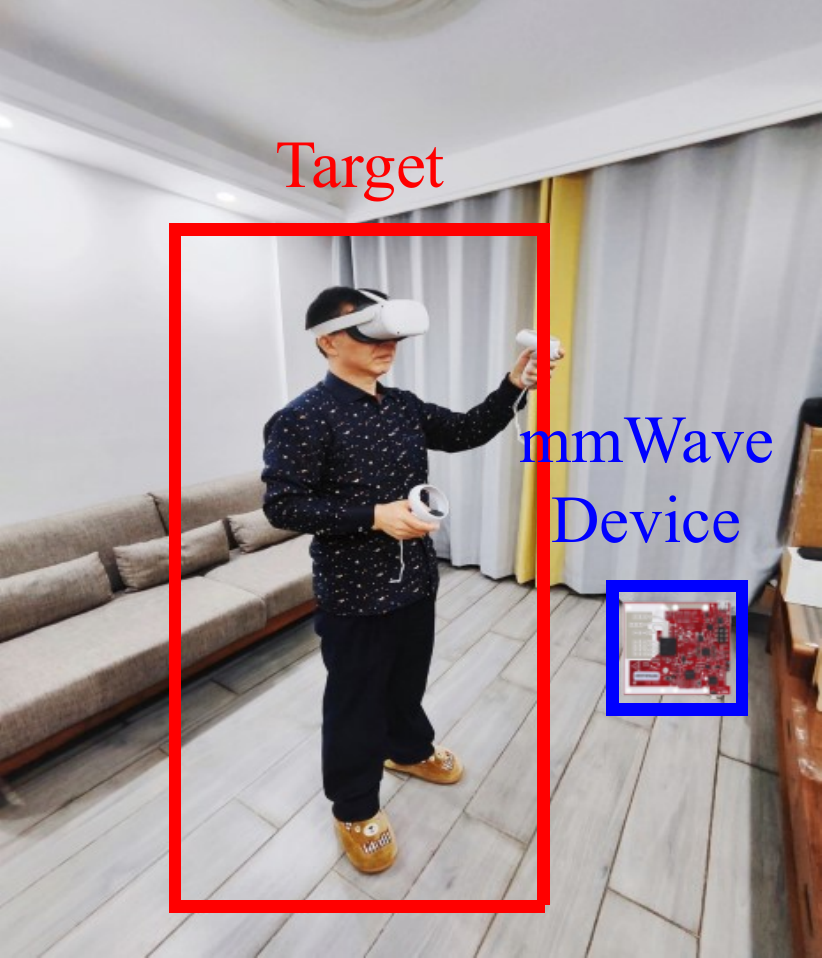}   
        \caption{The scenario involves a VR user and mmWave device.}
        \label{fig:attackdevice}
    \end{subfigure}
    \hspace{1mm}
    \begin{subfigure}[t]{0.46\linewidth}
        \includegraphics[width=\linewidth]{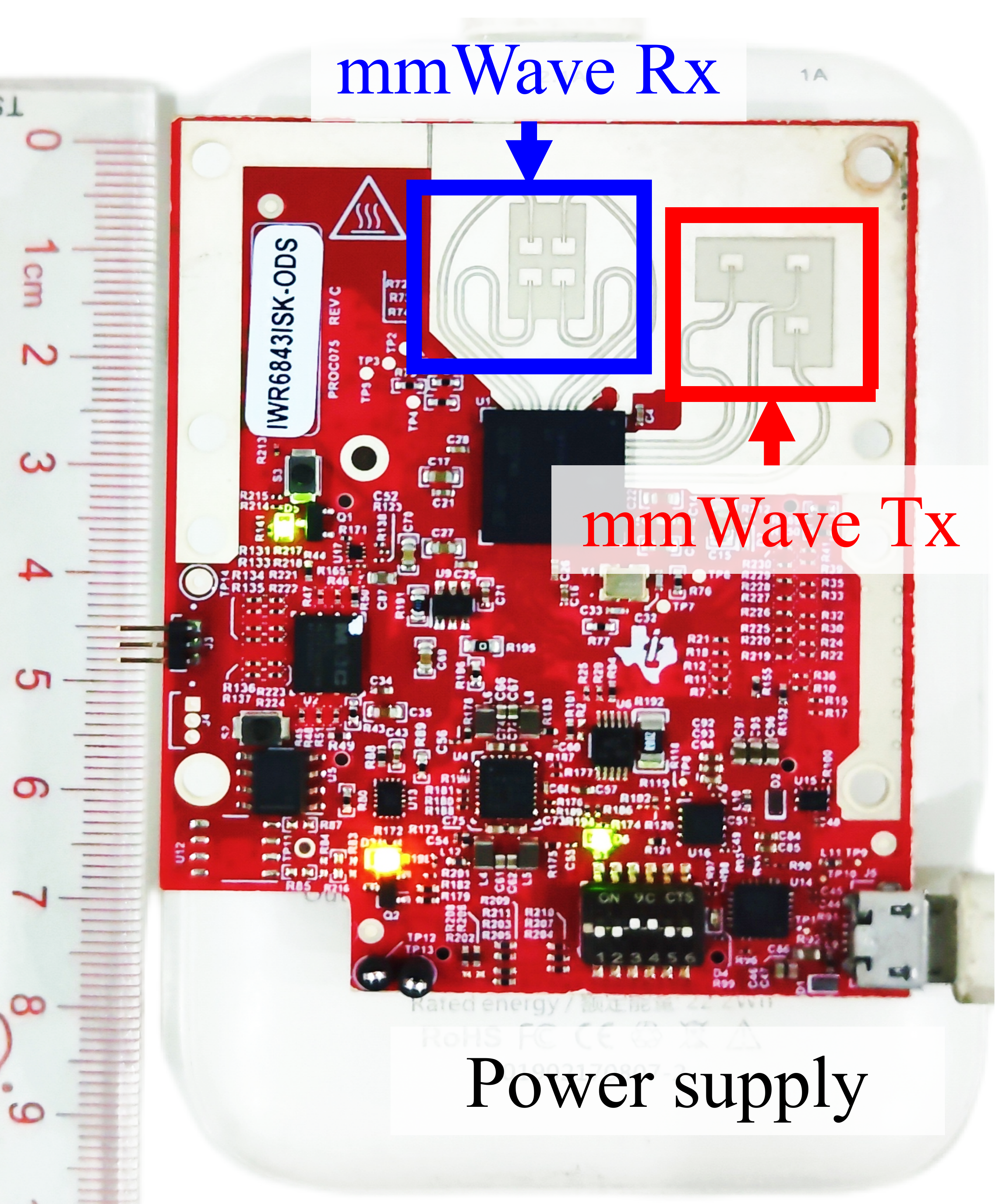}
        \caption{Detailed components in the mmWave device.}
        \label{fig:victimdevice}
    \end{subfigure}
\caption{ESP-PCT data collection scenario and platform.}
\label{fig:implementation}
\end{figure}

In this section, we present the data collection of the ESP-PCT prototype, which aims to obtain mmWave data for VR semantic recognition tasks. We designed the data collection components of the ESP-PCT to fit into a power bank, with a weight of 96 grams and dimensions of 8 cm in length and width. We utilize commercial VR devices as the target devices, as illustrated in Fig.~\ref{fig:implementation}. We recruited 12 participants, six males and six females, aged from 21 to 58, for our study. We obtain informed consent from each participant before the data collection. We collect 3,600 data sets comprising a 12TB dataset of mmWave point cloud and Kinect data, with 100 sets for each keystroke category. The mmWave radar data contained in each dataset include 30 seconds of raw signal, i.e., in-phase and quadrature components, and point cloud data, with a sampling rate of 10 frames. To the best of our knowledge, this is the first point cloud data set for VR semantic recognition, and we have made our dataset publicly available for future research and development.

\subsection{Implementation}
We evaluate ESP-PCT on the VR keystrokes recognition task, built on the 12TB dataset of VR semantics we collected. All our ESP-PCTs utilize a patch size 16×16 to partition the point clouds. The input of our model is a sequence of point clouds, each representing a frame of the VR user. We set the sequence length to 25, which means we utilize 25 frames to recognize one keystroke. All the training strategies, such as data augmentation, regularization, and optimization, strictly follow the original settings of ESP-PCT. We train ESP-PCT for a total of 700 epochs. To improve performance and prevent overfitting, we utilize early stopping with 200 epochs. We initialize the best validation loss to infinity and update it whenever a lower loss is found.

\subsection{Experimental Results}

In this section, we demonstrate comprehensive experiments on ESP-PCT within a range of real-world noisy VR environments, spanning no-occlusion to combined-occlusion scenarios and across point clouds of various densities. 

We evaluate ESP-PCT with the Baseline model~\cite{Generative_PointNet} and the state-of-the-art (SOTA) models including the Point Transformer~\cite{PointTransformer}, Point 4D Transformer~\cite{Point4DTransformer}, Stratified Point Transformer~\cite{StratifiedPCT}, and Self-Supervised4D~\cite{Self-Supervised4D}. These are represented by models focusing solely on preprocessing and extracting the semantic significance from point clouds. The evaluation metrics include Top-1 accuracy, FLOPs, and the number of parameters. The AppNet and KeyNet execute the final tasks of VR semantic recognition.

Table~\ref{tab:overallperformance} demonstrates that  ESP-PCT consistently surpasses other models in accuracy and efficiency for VR semantic recognition utilizing mmWave point clouds. In the no occlusion scenario, ESP-PCT achieved a top-1 accuracy of 97.6\% in application type recognition and 92.8\% in keystrokes recognition while requiring only 0.9 FLOPs(G) and utilizing 693 parameters (K). This superior performance extends to the wood occlusion scenario, with an accuracy of 94.1\% and 88.7\% for the respective tasks while maintaining low computational resource usage, a trend consistent across all tested scenarios. In brick occlusion experience, ESP-PCT achieved accuracies of 92.3\% and 86.9\%, showcasing its robustness. Even in the challenging combined occlusion scenario, ESP-PCT delivered accuracies of 89.4\% for application type recognition and 83.2\% for keystroke recognition, further demonstrating its capability to effectively recognize point clouds with intricate VR semantics while reducing computational and storage overheads simultaneously.

The results of the ablation study for the ESP-PCT model presented in Table~\ref{tab:ablationperformance}, highlight the crucial role played by four key components: attention scores, group scores, the highest sum, and Top-K points. This study demonstrates how each element contributes to the model's high accuracy and computational and memory efficiency. By honing in on regions of the point cloud that are semantically discriminative, ESP-PCT effectively narrows its focus to segments teeming with VR user action semantics. This targeted approach decreases the volume of data that needs to be processed and enhances the model's accuracy and overall efficiency. When all the designed features of ESP-PCT are combined, the model attains peak accuracy and computational economy performance. The implementation of this approach indicates that by harnessing the rich semantic information extracted from the point cloud data associated with VR user movements, the model effectively discards redundant data, thereby boosting accuracy and cutting down on unnecessary complexity.

\begin{figure}[h]
    \centering
    \includegraphics[width=1\linewidth]{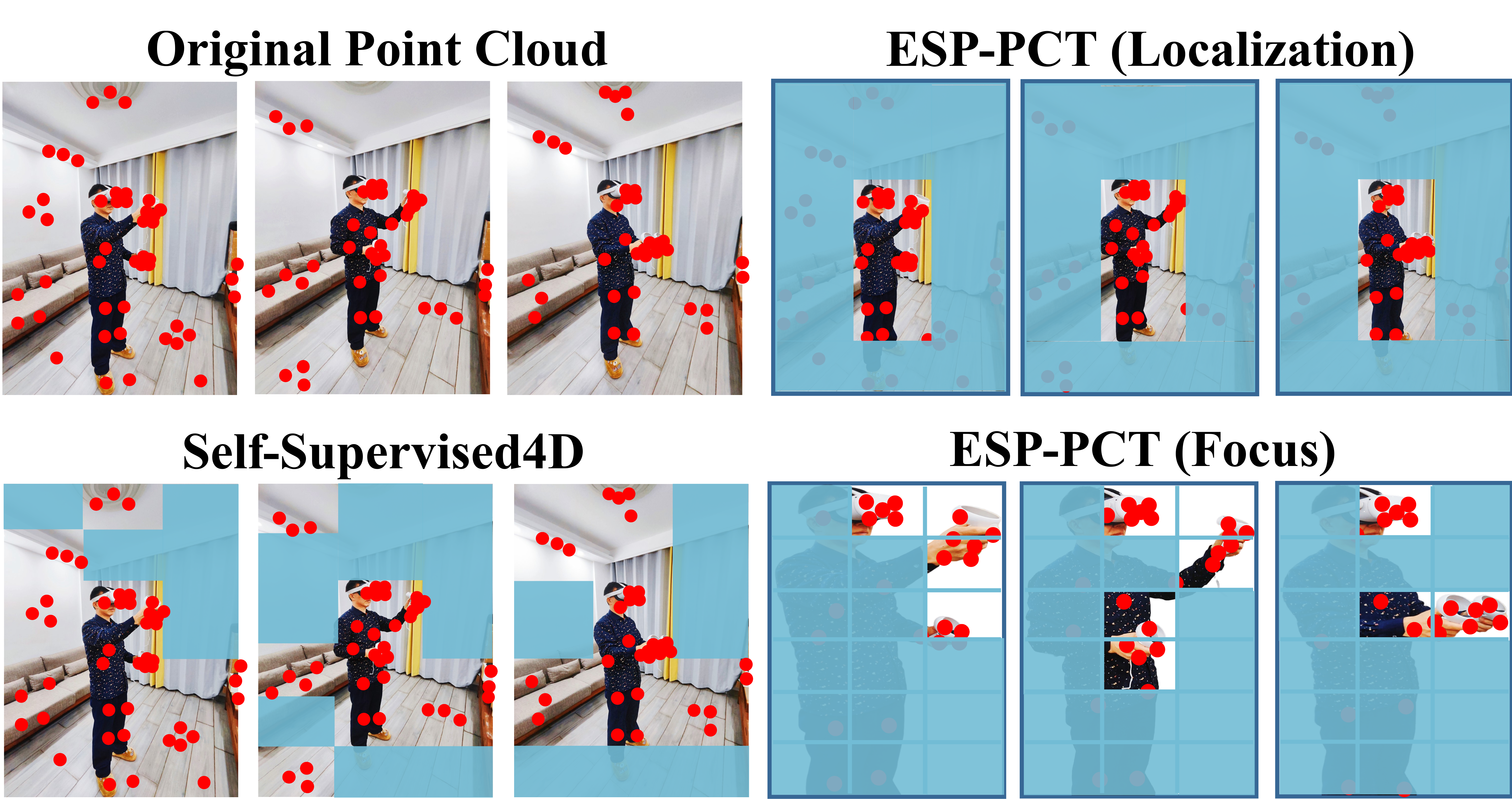}
    \caption{Semantic-discriminative region selected by ESP-PCT, Self-Supervised4D, and original point could.}
    \label{fig:focuscomparison}
\end{figure}

Fig.~\ref{fig:focuscomparison} presents a comparative analysis of the point cloud regions selected by our ESP-PCT during localization and focus stages against those chosen by Self-Supervised4D~\cite{Self-Supervised4D}. This figure comprises four sets of three subfigures, each depicting successive actions by a VR user. The mmWave point clouds are showcased at the top left, with Self-Supervised4D's selections at the bottom left and ESP-PCT's chosen regions for localization and focus at the top right and bottom right, respectively. The comparative visualizations highlight that Self-Supervised4D is susceptible to environmental noise and often fails to isolate point cloud regions critical to VR semantics. In contrast, ESP-PCT proficiently identifies VR user-related point clouds by localizing and focusing on areas with dense VR semantic content.

\begin{figure}[t]
\centering
    \begin{subfigure}[t]{0.46\linewidth}
        \includegraphics[width=\linewidth]{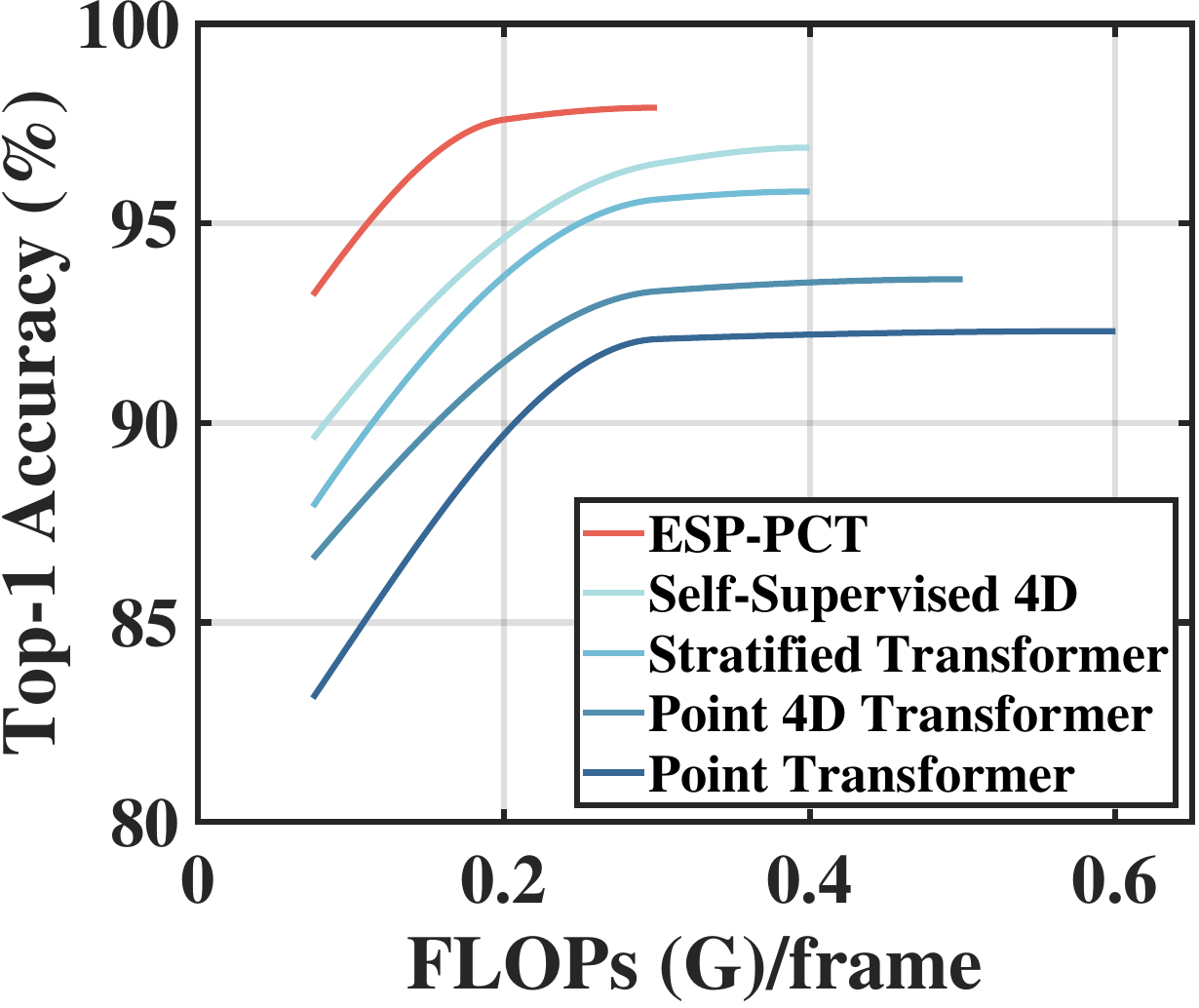}   
        \caption{Application Recognition.}
        \label{fig:appcomparison}
    \end{subfigure}
    \hspace{1mm}
    \begin{subfigure}[t]{0.46\linewidth}
        \includegraphics[width=\linewidth]{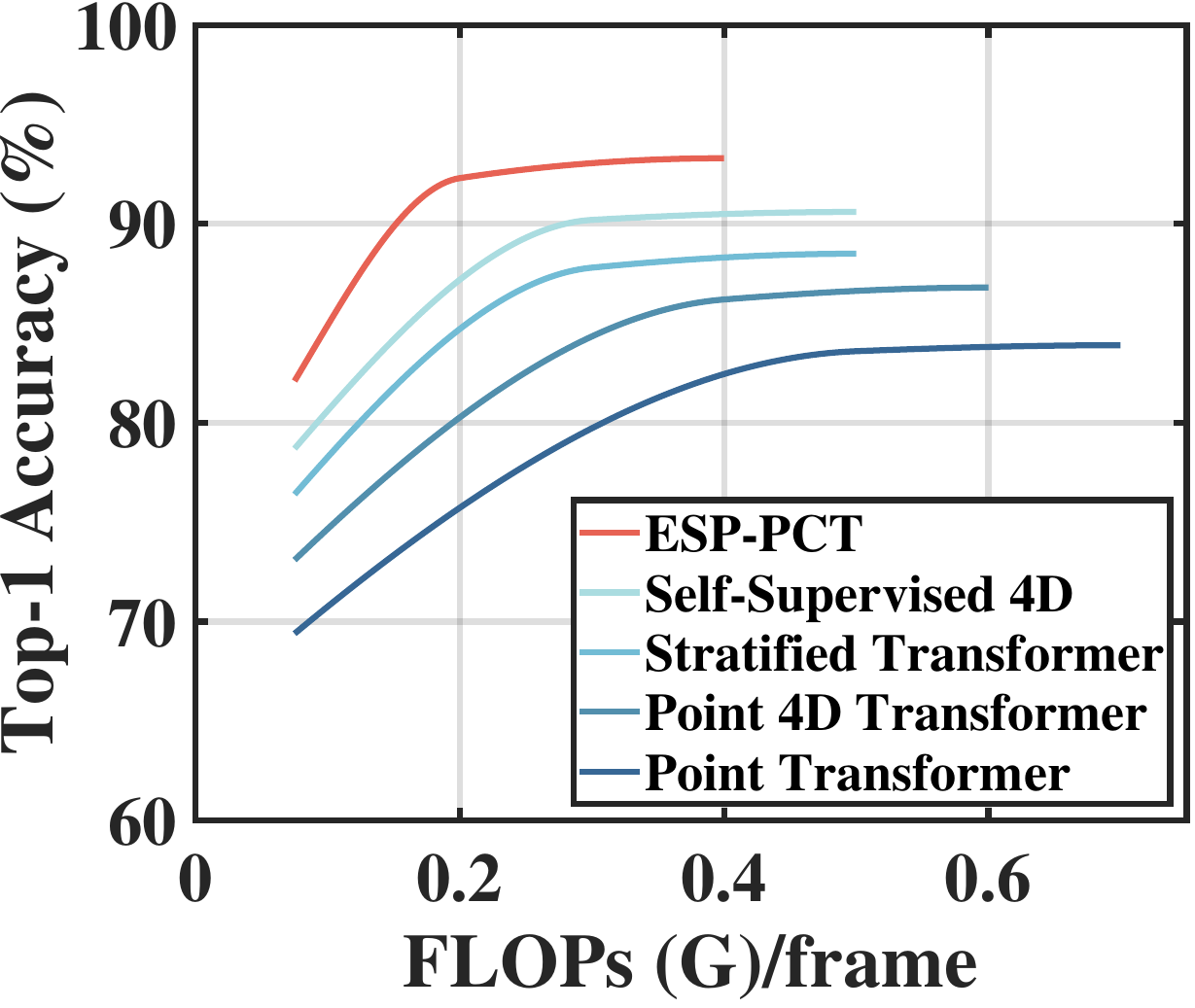}
        \caption{Keystrokes Recognition.}
        \label{fig:keycomparison}
    \end{subfigure}
\caption{Comparison of semantic recognition accuracy between ESP-PCT and SOTA approaches in application recognition and keystroke recognition tasks.}
\label{fig:comparison}
\end{figure}

We conduct comparative experiments on our ESP-PCT with SOTA methods with application type recognition and keystroke recognition. Our ESP-PCT outperforms other models in accuracy and efficiency, as shown in Fig.~\ref{fig:appcomparison} and Fig.~\ref{fig:keycomparison}. This advantage arises from ESP-PCT's proficiency in managing the data sparsity challenge inherent in mmWave point clouds. Despite reducing FLOPs (G)/frame complicating the extraction of information, ESP-PCT preserves VR semantics points and eliminates environmental noise. In contrast, SOTA models retain environmental noise, compromising their ability to discern VR semantic-related features and ultimately degrading their accuracy. Moreover, ESP-PCT reduces computational overhead by 76.9\% compared to the Point Transformer~\cite{PointTransformer}, resulting in an estimated inference time of 35ms, ensuring real-time processing within the typical mmWave radar sampling interval (10 Hz).

Specifically, for application type recognition, ESP-PCT achieves a 97.9\% accuracy at 0.3G/frame in FLOPs, marking a significant 5.6\% improvement over the Stratified Transformer's performance. For keystroke recognition, which includes a larger number of categories (26 letters and 10 digits), ESP-PCT outperforms with a top-1 accuracy of 93.3\% at 0.4 G/frame in FLOPs, which is an impressive 10.8\% higher than the Point Transformer’s performance. Furthermore, ESP-PCT demonstrates superior efficiency in computational cost relative to other methods. For instance, at a 95\% accuracy level for application type recognition, ESP-PCT requires 0.2G/frame in FLOPs, which is one-third of the Self-Supervised 4D’s requirement and significantly lower than other models. In terms of keystroke recognition targeting a 90\% accuracy level, ESP-PCT’s computational demand is 0.2 G/frame in FLOPs, which is half of the consumption of Self-Supervised 4D.

%% file: sec5_conclusion.tex
\section{Conclusion}
\label{sec6_conclusion}

This paper presents ESP-PCT, a novel framework that improves the model accuracy while reducing redundancy by dynamically locating and focusing on the semantic-discriminative regions in the point cloud data generated from mmWave signals. The key insight is that not all points in a point cloud are equally important, empowering the framework to process data selectively and emphasizing the most informative regions to enhance semantic analysis. We validate the effectiveness and efficiency of ESP-PCT on various VR semantic recognition tasks utilizing point cloud data, and we release a 12TB dataset of mmWave point cloud and Kinect data under various VR scenarios for further research. Future work could explore applying ESP-PCT to other objects and scenarios for semantic recognition.